\documentclass{IEEEtran}
\usepackage[utf8]{inputenc}
\usepackage{tikz}
\usepackage{smartdiagram}
\usepackage{multirow}
\usepackage{balance}

\title{Quality Monitoring and Assessment of Deployed Deep Learning Models for Network AIOps}
\author{Lixuan Yang and Dario Rossi -- Huawei Technologies France SASU,  \texttt{first.last@huawei.com}
}
\usepackage{graphicx}
\usetikzlibrary{arrows}
\usepackage{xcolor}

\usepackage{cite}
\usepackage{footnote}
\usepackage{hyperref}
\hypersetup{
    colorlinks=true,
    linkcolor=blue,
    filecolor=magenta,      
}

\usepackage{cleveref}
\usepackage{booktabs}
\newcommand{\DR}[1]{\textcolor{red}{#1}}
\newcommand{\chomp}[1]{}
\usepackage{comment}

\begin{document}
\bstctlcite{IEEEexample:BSTcontrol}

\maketitle

\begin{abstract}
Artificial Intelligence (AI) has recently attracted a lot of attention, transitioning from research labs to a wide range of successful deployments in many fields, which is particularly true for Deep Learning (DL)
techniques.
Ultimately, DL models being software artifacts, they need to be regularly maintained and updated: AIOps is the logical extension of the DevOps software development practices to AI-software applied to network operation and management.
In the  lifecycle of a DL model deployment, it is important to assess the quality of deployed models, to detect ``stale'' models and  prioritize their update. In this article, we cover the issue in the context of network management, proposing simple yet effective techniques for (i) quality assessment of individual inference,  and for (ii) overall model quality tracking over multiple inferences, that we apply to two use cases, representative of the network management and image recognition fields.
\end{abstract}

\section{Introduction}

A decade ago, Marc Andreessen was rightly anticipating that ``software is eating the world'': Artificial Intelligence (AI) software has undoubtedly  the very same appetite for the next decade.  
In particular, spectacular successes of Deep Learning (DL) techniques have gained significant share of newspaper attention in the image classification field, with numerous models already deployed in production.
In a nutshell, these DL advances have equally benefited  from ground-breaking theoretic research\footnote{Y. Bengio, Y. LeCun and G. Hinton, 2018 ACM Turing award}, from  software plaforms efficiently exploiting hardware acceleration\footnote{J. L. Hennesey and D. Patterson, 2017 ACM Turing award},  
and from the availability of large \emph{corpuses} of labeled data such as ImageNet (comprising tens of millions of images for tens of thousands of classes). 

Without willing to undermine the significance of DL achievements, manifest challenges in replicating such corpuses for other domains, and for networking in particular, are rooted in (i) the lack of standard data representations, as well as in  (ii) the pace of technology evolution, which is much faster with respect to the evolution of the physical world.  Indeed (i) data in network field is significantly more heterogeneous and diverse than the well normalized image format. Additionally, (ii) while training a DL model to recognize cats requires a significant volume of labeled  cat images, however the \emph{Felis}  genus, which includes the domestic cat, has not evolved in the last six million years. This is in stark contrast
with the pace of network technology evolution,  well captured by Gilder law observing that \emph{``Bandwidth grows at least three times faster than the compute power''} -- i.e., even faster than the already exponential growth in the computing capacity postulated in the widely popular Moore law.

The  network research community, as many other fields,  is exploring the use of DL models to relieve and assist human operation for diverse tasks, such as dynamic resource management~\cite{9083670}, traffic classification and management~\cite{9146411} and troubleshooting.
Yet,  the network fast paced evolution, combined to the lack of a universal network data format and privacy/business-sensitive matters, have so far made it impossible to construct and share large corpuses of networking data, which were crucial for DL success in the image field.  This yields to several challenges for DL deployment in the network field. First,  the knowledge gained by DL models during training is hard to transfer, thus special attention should be payed to assess whether the model is fit to the environmental conditions where it has been deployed: e.g., a DL model for resource management~\cite{9083670} trained on outdoor settings is hardly applicable to indoor ones.  
Second,  as per the just observed fast-paced evolution of the  network traffic and technological  landscape,  it is reasonable to expect  that any deployed DL model should be regularly updated, to remain fit to the job it has been originally designed for: for instance, a DL model for traffic classification~\cite{9146411} will encounter new applications it was not trained to recognize. Therefore, for successful application of DL in the  networking domain, there is need for lightweight techniques capable of (i) \emph{assessing the quality of each inference} of already deployed models, as well as (ii) \emph{tracking the quality over multiple inferences} of the same model: such techniques would allow  to timely prioritize and trigger model updates, guaranteeing that deployed models remain fit for its task even in a changing environment.

Given these premises, this work addresses the  challenges of DL model monitoring in the network O\&M context. We start by introducing the lifecycle of a DL model (\S\ref{sec:aiops}), contextualizing the challenges  to evaluate the quality of deployed DL models in the popular AIOps framework (\S\ref{sec:assess}). Particularly, we identify Open Set Recognition (OSR) techniques as a key building block for quality assessment and tracking, and we contextualize our proposed techniques in the context of related OSR work in (\S\ref{sec:related}).   We finally introduce two example use-cases, pertaining to the computer vision and network O\&M fields respectively, where we apply several of the aforementioned OSR techniques for DL quality assessment and tracking (\S\ref{sec:use-case}) and summarize the key points (\S\ref{sec:conclusion}).

\section{AIOps: the broad picture}\label{sec:aiops}

\newcommand{\smallcaption}[1]{\begin{scriptsize}\flushleft #1\end{scriptsize}}

\begin{figure}
    \centering
    \includegraphics[width=\columnwidth]{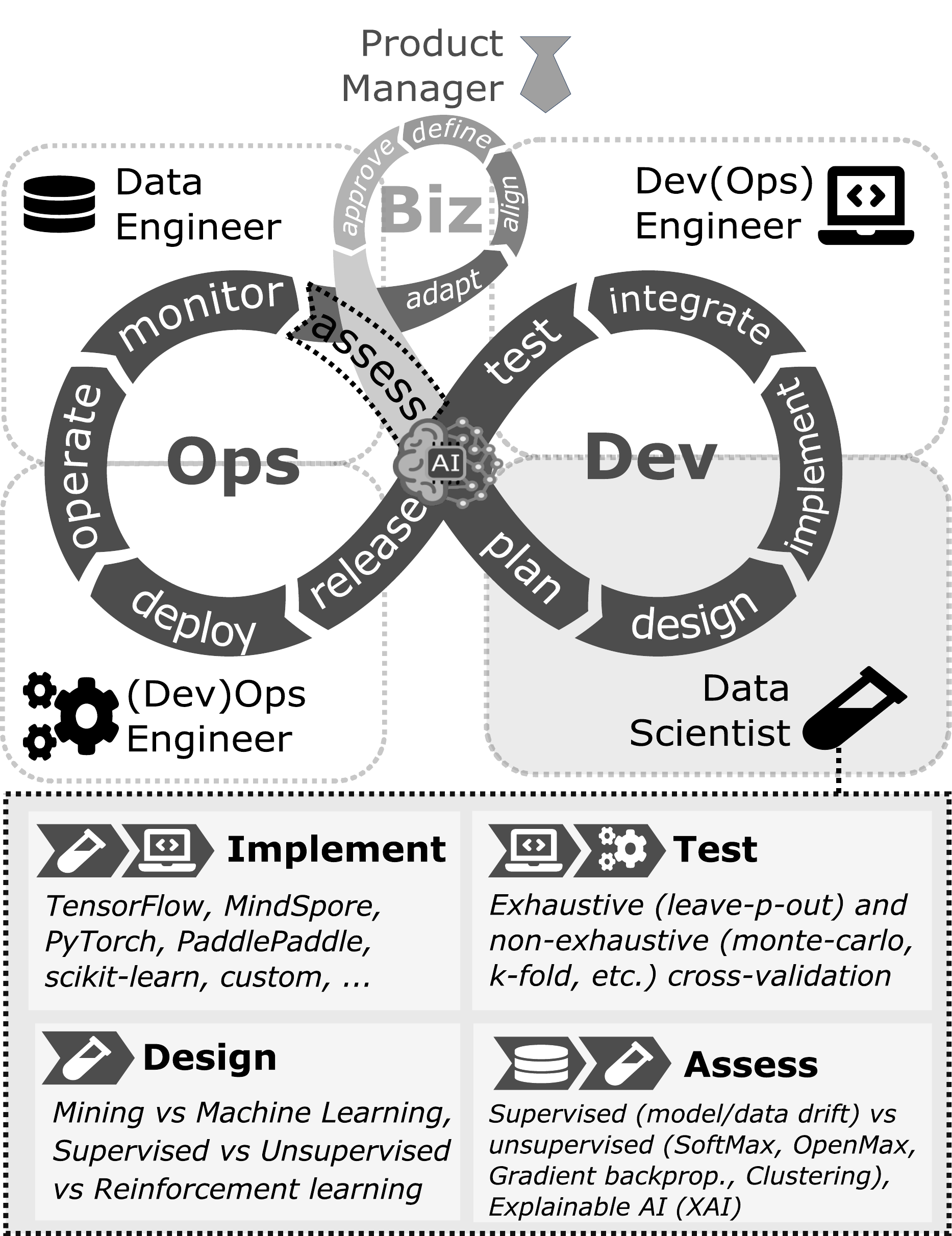}
    \caption{\textit{Synoptic of Network AI workflow}: (top)
     main actors in the BizDevOps agile method, with special focus on the AIOps roles and interactions;
    (bottom)  high-level taxonomy of data science tools, practices and techniques useful at different stages of the BizDevOps loop. While data scientists have full control over the \emph{design} of the AI solution,  data science methods are still useful in several steps of the workflow (e.g., \emph{implement, test, assessment}) that are out of the hands of the data scientist (e.g., under control of developer, IT operator and data engineer respectively).
    Crucially, the BizDevOps and DevOps methodologies fork after the monitoring of deployed models (\emph{adapt} vs \emph{assess}): however, data engineer and product manager profiles lack data science skills to interpret the monitored data, hence weakening the alignment decisions. In this work, we focus on  data science techniques for automated model \emph{assessment},  empowering  Biz deciders to take data-driven decisions about prioritization of data-engineering effort (e.g., labeling, data quality, etc.),  ameliorating cross-team interaction and the AIOps loop effectiveness overall.}
    \label{fig:aiops}
\end{figure}

AIOps is a catchy term recently coined by Gartner to identify trends in deployment of AI techniques in a network O\&M context using an agile development methodology.  We introduce the network AIOps workflow through the synoptic illustrated in \Cref{fig:aiops}.

\subsection{Agile methodologies}
AIOps in itself is an extension of DevOps, which applies agile methodologies to combines software development (Dev) and IT operations (Ops), aimed at shortening the systems development life cycle and provided continuous delivery with high software quality. 

Whereas prior to DevOps the lifecycle of new product releases was measured in months, DevOps shorten the duration to about two weeks.  In particular, DevOps works in sprint where planning and implementation phases (Dev) are followed by quick deployment in operational context (Ops): monitoring the output of the Ops phase serves as input to the next Dev phase, to iteratively improve the overall solution. 
Business (Biz) aspects can be additionally integrated in an agile manner by an extended BizDevOps loop: alignment with business objectives is done immediately after the Ops monitoring step, to adapt and define the next Dev phase.

\subsection{Agile + Network + AI}
AIOps is a particularization of the BizDevOps cycle to take into account specific characteristics of AI-software development. 
Complementary to the Dev and Ops engineering teams in classic IT and O\&M scenarios, AI development requires additional skillsets, which are identified in the Data Engineering and Data Scientist roles respectively. As the teams and the set of skills  grow, the interaction in the workflow becomes richer, but also more complex.

Far from providing exhaustive background on  AIOps, our goal is rather  focus on the \emph{data science} skills,  and in particular to identify which of the AIOps steps that are not directly owned by a  data scientist, could nevertheless benefit from a data science approach. Such steps would therefore benefit from transfer of  data science  techniques and practices, that can hopefully work with as little data scientist intervention as possible.

\subsubsection{Design}
AI design is the primary role of the Data Scientist, in accordance with business requirements as input. 
It is out of the scope of this article to provide a full background on the usual data scientist ``weapons'',  whose selection depends on the problem at hand, and that broadly pertain to the data mining  and machine learning fields. The bottom part of Fig.\ref{fig:aiops} reports a succinct (as these areas are well overviewed in the literature) taxonomy of such techniques, including but not limited to popular  DL techniques.
In particular, most of the successful and well known DL models are subject to a supervised training phase,  in which the DL model is presented with relevant and abundant examples of the objects to recognize (classification) or the real-valued function to learn (regression). With respect to AIOps, understanding if and when deployed DL models need to be retrained is thus an important task.

\subsubsection{Implement}
Implementation and integration are  the primary role of Dev Engineers.
To make a smooth transition from proof-of-concept development to integration into products, there exists a well established set of DevOps toolings -- from an AI perspective, it sufficient to agree on the AI software platform and allowed libraries, using best practices  to simplify replicability  (i.e., use of containers or well-defined library environments).  Clearly, in some cases custom implementation are required to comply with further product constraints, which ultimately can lead to alter the PoC DL models.

\subsubsection{Test}
Testing prior to deployment is an essential tasks of Dev Engineers: as DL models, or the data pipeline, or the data itself  may evolve during integration with respect to the design phase, testing guarantees the code to be ready for release and operational deployment. 
In this case, data scientists can contribute by sharing 
best practices, which are the AI-flavored version of 
classic unit/regression testing in the more general software
industry: one such practice is the use of  cross-validation techniques, specifying the way in which the training/validation/testing data should be partitioned  to ensure statistical relevance of the results,  avoiding overfitting  and biases.

\subsubsection{Assess}
After models are deployed by Ops engineer in production system, the Data engineers \emph{monitor} data from deployed DL models.
In the BizDevOps loop, the monitored data is used to adapt, align, define and agree the next Dev phase -- interposing between the Ops and Dev phases.  At the same time, it is easy to observe that Data Engineer and Product managers lack the necessary data science skills to gather profound observations from the monitored data.

The missing yet necessary link in the BizDevOps loop is represented by the \emph{assessment} of the quality of deployed DL models. More precisely, we argue that there is need for \emph{automated assessment} techniques, so that the monitoring output can be automatically enriched with data science analytics, e.g., allowing to understand if models are still fit for the environment where they have been deployed, or if they have become ``stale'' and need retraining.
Such quality assessment can apply (i) to \emph{individual inferences}, enriching the DL output with third-party confidence about its accuracy, which can be useful to   decide whether to trust \emph{punctual decisions} of a DL model, e.g., a traffic classification result~\cite{9146411}. 
Moreover, quality  assessment can span (ii) over \emph{multiple inferences}, which allows \emph{tracking the overall quality} of a  DL model over time, e.g., which can be useful to trigger model retraining, or model switching, if environmental conditions change~\cite{9083670}.

Such techniques, on which we focus on the remainder of this paper, are crucial to ensure that business alignment is taken on the ground of objective, accurate, significant and easily interpretable output. Particularly, in what follows we propose, contrast and experimentally evaluate techniques that allow to (i) assess the  quality of individual inferences, as well as (ii) tracking the overall model quality.

\section{Quality assessment of DL models in AIOps}\label{sec:assess}

\subsection{Quality assessment challenges in network O\&M} 
Quality assessment methods need to address a number of network-specific challenges, that we now overview.

\subsubsection{Label cost}
Depending on the task, gathering new labels can either  be a relatively simple or a very complex operation. For instance, gathering a stream of labels is possible for (i) \emph{forecasting} (e.g., predicting the future value based on previous inputs) and \emph{regression} tasks (e.g., predicting the current value of an observable based on other inputs). In such cases, the current stacks supporting AIOps operations such as MLFlow (\url{https://mlflow.org/}) and MindSpore (\url{https://github.com/mindspore-ai}) are already well suited for continuous learning,  combating model ``drift'' by seamlessly integrating   data collection and model update.
Conversely, label can be costly to obtain for  (ii) \emph{classification} tasks: in such scenarios, a Biz decision ought to be made concerning which of the deployed models are in more urgent need for further labeling, for which a measure of models ``staleness'' would be of significant help. Open-set recognition (OSR) techniques developed in the data science community~\cite{NDsurvey} are a good fit for this: shortly, OSR approaches are aware that DL models are trained with an incomplete view of the world (a finite set of ``known'' classes), and defines way to effectively deal with previously unseen classes (or ``unknown'' at training time) that can be submitted to an algorithm in real deployment. OSR capabilities are thus crucial in AIOps, as for instance for the case of application identification~\cite{9146411} that we consider as illustrative example in the experimental evaluation   (\S\ref{sec:use-case}).

\subsubsection{Training risk}
Whereas models can be incrementally trained, the quality of the labeling process affects the model quality. For instance,  the natural class imbalance present in the data can bias the models and lead to unfairness/low performance unless properly accounted for -- e.g., by stratified input sampling, or by using appropriate loss functions in the DL models. Similarly,   samples can be mislabeled (inadvertently, or even on purpose) which is referred to as adversarial training, and can greatly confuse the model.   The aforementioned OSR techniques can assess DL model staleness with respect to a new stream of labels:  roughly quantifying the magnitude of changes that models should undergo to properly account for the new labels, could be useful to both (i) \emph{assess the retraining cost},  and also assist ameliorating the (ii)  \emph{quality of the labeling} process (e.g., suspicious label detection).
In what follows, we also introduce techniques allowing to timely detect 
model quality degradation, even from the first weak signals (\S\ref{sec:use-case:tracking}).

\subsubsection{Interpretability}
Explainability of AI methods~\cite{XAIsurvey}, also known as XAI, has been identified as crucial for AI application  in  many domains. While issues tied to fairness and ethical concerns are not predominant in the  typical set of network O\&M tasks, however a set of management-related aspects (such as   auditability,  accountability,  legal and forensic matters, etc.) would greatly benefit from
model interpretability.  
Similarly to individual model output/decisions, assessment of DL model quality should abid the \emph{same  interpretability criteria}, to simplify interaction with non-AI experts (i.e., Ops and Biz primarily): in turn, combining OSR and XAI for accurate and interpretable DL model quality assessment would allow to better prioritize business decisions, filling a crucially missing link of the BizDevOps loop.


\subsubsection{Deployability}
From a practical viewpoint, it is desirable for DL quality assessment techniques to work on (i) \emph{pre-existing and unmodified} models: the need to alter DL models, by e.g., using a specific DL architecture, could negatively affect adoption of the quality assessment techniques, as it would entail complete retraining of all deployed DL models -- at a likely prohibitive cost.
Similarly, it is reasonable to assume that quality assessment techniques can access existing models, their input and outputs, but should otherwise (ii) \emph{rely on  the least possible amount of meta-information about the model} -- for instance, access to training data may be very hard in practice.
Additionally, corpuses in the image field offer many examples of unlikely inputs: e.g., a software processing X-ray images in medical field will unlikely have to process images of cats, cars and seaside sunsets. As such, image corpuses offer plenty of instances of the so-called Out-of-Distribution (OOD) samples that OSR techniques are designed to detect. Conversely, it is harder to provide these examples in the networking field. This is due, on the one hand, to the aforementioned lack of corpuses and, on the other hand, to the evolutionary nature of the network traffic and technology: for instance, as new applications are developed every year, traffic  identification engine\cite{9146411} will by design be confronted with zero-day applications. As such, OSR techniques for networking use-cases should not make any assumption on OOD data, and do not rely on OOD data for fine-tuning, which instead  the common-place in the image field.
Finally, (iii) \emph{flexibility} to assess  quality of  individual model inferences, as well as ability to track the overall model quality over time, are important practical features.

\usetikzlibrary{arrows,shadows,positioning}

\tikzset{
  frame/.style={
    rectangle, draw, 
    text width=6em, text centered,
    minimum height=4em,drop shadow,fill=yellow!40,
    rounded corners,
  },
  line/.style={
    draw, -latex',rounded corners=3mm,
  }
}

\begin{figure}
\centering
\resizebox{\columnwidth}{!}{%

\begin{tikzpicture}[font=\small\sffamily\bfseries,very thick,node distance = 2cm]

\node[inner sep=0pt] (cnn) at (0,0)
    {\includegraphics[width=\textwidth]{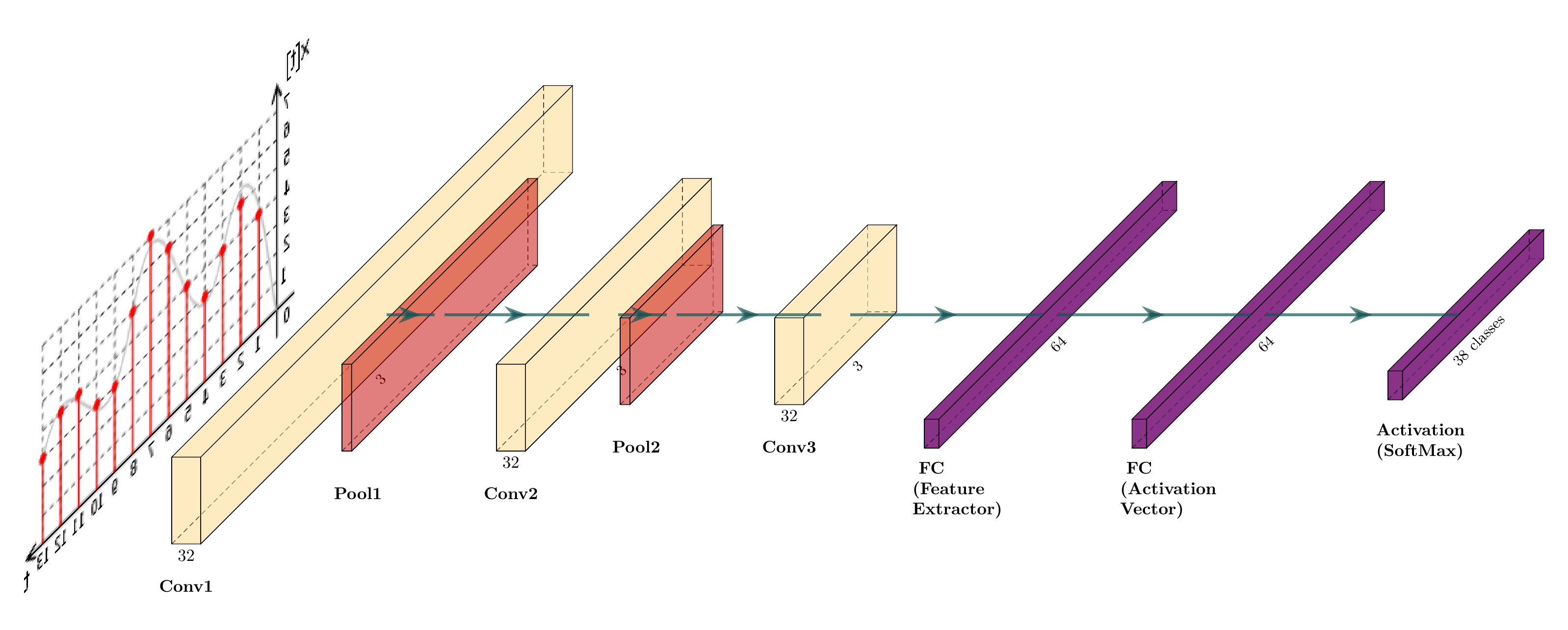}};
\node[frame] (C) at (-5,4) [draw,thick,minimum width=1cm,minimum height=1cm] {\large Input Preprocessing \cite{Zhang15,Liang17,Lee18}};

\node[frame] (D) at (2,4) [draw,thick,minimum width=1cm,minimum height=1cm]   {\large Loss\\Function\cite{Hassen18}};
\node[frame] (F) at (4,-5) [draw,thick,minimum width=1cm,minimum height=1cm] {\large Classifier \\ K+1 \cite{Neal2018}};
\node[frame] (K) at (5,4) [draw,thick,minimum width=1cm,minimum height=1cm] {\large Softmax \cite{Hendrycks2017, Liang17}};

\node[frame] (B) at (-3,-3.8) [draw,thick,minimum width=1cm,minimum height=1cm] {\large Autoencoder};
\node[frame] (A) at (0,-5) [draw,thick,minimum width=1cm,minimum height=1cm] {\large GAN};

\node[frame] (L) at (0,-3) [draw,thick,minimum width=1cm,minimum height=1cm] {\large Reconstruction Error}; 

\node[frame] (I) at (6,-2) [draw,thick,minimum width=0.2cm,minimum height=1cm] {\large OpenMax\\ \cite{Bendale15}};

\node[frame] (J) at (6,-4) [draw,thick,minimum width=1cm,minimum height=1cm] {\large Mahalanobis\cite{Lee18} \\ Clustering\cite{Zhang20}};
\node[frame] (G) at (8,4) [draw,thick,minimum width=1cm,minimum height=1cm] {\large Gradient\\based\cite{Yang2021}};

\path [line] (C) -- node[right,align=left,pos=.5] {\large Augment\\\large Input}(-5, 1);
\path [line] (-5, -2) |- (B) ;
\path [line] (B) |- (A) ;
\path [line] (D.350) -- (7, 0) ;
\path [line] (G) -- (7, 0) ;
\path [line] (K) -- (7, 0) ;
\path [line] (B) |- node[right,align=right,pos=.7] {\large Latent \\[3mm]\large Repr.} (L) ;
\path [line] (A) -- node[right,align=left,pos=.1] {\large Generate \\[3mm]\large Unknowns} (F) ;
\path [line] (J) -- (2, -1) ;
\path [line] (I) -- (4, -1) ;

\end{tikzpicture}
} 

\vspace{12pt}

\begin{footnotesize}
\begin{tabular}{llll}
\toprule
& {\bf (Ref.) Technique} & {\bf Deployment} &  {\bf Complexity}  \\ 
\midrule 

\multirow{3}{*}{\rotatebox{90}{{\bf In.}}}
&\chomp{2015}\cite{Zhang15}   Input clustering & Unmodified ML/DL &  High\\
&\chomp{2017} \cite{Liang17}  Temperature scaling & Unmodified DL &  High \\
&\chomp{2018} \cite{Lee18}    Mahalanobis distance & Unmodified DL & High \\

\midrule 
\multirow{2}{*}{\rotatebox{90}{{\bf Arch.}}}
&\chomp{2018} \cite{Hassen18}  Clustering loss & Modified ML/DL&  Low \\  
&\chomp{2018} \cite{Neal2018}  K+1 Classifier & Modified ML/DL & Low\chomp{(threshold)}\\


\midrule 
\multirow{4}{*}{\rotatebox{90}{{\bf  Out.}}}
&\chomp{2017} \cite{Hendrycks2017}  SoftMax & Unmodified ML/DL & Low\chomp{ (threshold)}\\
&\chomp{2015} \cite{Bendale15}   OpenMax (AV) & Unmodified ML/DL &  Medium\chomp{(Weibull inference)} \\
&\chomp{2020} \cite{Zhang20}  Clustering (FV)& Unmodified DL &  High\chomp{ (clustering)} \\
&\chomp{2021} \cite{Yang2021}  Gradient-based& Unmodified DL &  Low\chomp{ (gradient)} \\

\bottomrule
\end{tabular}
\end{footnotesize}

\caption{\emph{State of the art}: Overview and taxonomy of OSR techniques, useful for  quality assessement of DL model individual inferences.  Acronyms: Feature Vector (FV); Activation Vector (AV); Autoencoders (AE), Generative adversarial networks (GAN). }
\vspace{-15pt}
\label{fig:related}
\end{figure}

\subsubsection{Efficiency and timeliness}
From a (i) \emph{computational complexity}  viewpoint, it appears necessary for  DL model quality assessment to be a relatively simple operation~\cite{9083670}. 
Ideally, quality assessment of a model output should have a cost comparable to the inference cost of the same model -- which would allow to track the quality of deployed models at a fine grain. Similarly, quality assessment operations should be significantly much simpler with respect to model training -- as otherwise, quality assessment step could be safely skipped, to directly plan a fully-fledged incremental training instead.
Finally, a related but complementary  property concerns the (ii) \emph{timeliness} of the model quality detection: in other terms, irrespectively of the cost of individual assessment operation for a specific input/output pair, in some mission-critical cases it would be desirable to have techniques able to immediately detect model drifts from the very first few samples.

\subsection{Quality assessment state-of-the-art} \label{sec:related}
Solutions for DL quality assessment able to overcome the above challenges
can be cast to the more specific problem of 
(i) evaluating the adequacy of a trained model to handle the typical input of the environment it is deployed to and (ii) doing that in a flexible, deployable, interpretable and efficient manner.  These tasks can be achieved by building on  recent work pertain to the OSR~\cite{NDsurvey}  and  XAI~\cite{XAIsurvey} fields respectively. With respect to~\cite{NDsurvey},  Fig.~\ref{fig:related}  focuses on more recent work that exploits latest advances in DL, and that tackles quality assessment at the DL input, architecture or  output stages.

\subsubsection{Input}
A first set  of work tackles the problem directly in the input space, complementing the supervised DL models building a clustering representation of the input space: the further input samples are from all clusters, the less fit is the model for this data.
A key challenge when using input-clustering is the creation of clusters fitting well the different classes: to better control this, instead of working directly on the original input space~\cite{Zhang15} (which is independent of the DL model and thus inherently non-explainable),  some proposals apply transformations on the input so control models output, e.g. by mean of temperature-scaled~\cite{Liang17} or Mahalanobis~\cite{Lee18} scores (that are closer to the model and thus better interpretable).   However, the main drawback of these OSR proposals is their additional computational cost tied to the evaluation of the distance of new samples from all clusters -- which makes  model quality assessment rather complex.  

 \subsubsection{Architecture}
At their core, DL methods project input data into a \emph{latent space} where is easier to separate data based on class labels: a set of work proposes specific ways to alter  this  latent space to purposely simplify OSR task.   For instance,  ~\cite{Hassen18} use clustering loss functions to  constraint points of the same class to be close to each other,  to simplify the clustering task. Other work \cite{Neal2018} instead uses Generative Adversarial Networks (GAN) to  generate 
counterfactual samples of a fictitious ``unknown K+1 class''. Methods in this branch require specific architectures  and  extra training -- while possibly relevant for new deployments,  they  do not apply to existing ones, and hence have low practical appeal.


\subsubsection{Output}
The most interesting class of OSR approaches leverages model output.  The basic approach is thresholding SoftMax outputs~\cite{Hendrycks2017}, which simple and directly explainable, yet is not robust to overconfidence of the DL models.  To counter this problem, OpenMax~\cite{Bendale15} revises SoftMax activation vectors adding a special ``synthetic'' unknown class (with weigths induced by Weibull modeling).  To overcome curse of dimensionality, ~\cite{Zhang20} performs clustering at the output of the neural network, with a PCA dimensionality reduction.  In~\cite{Yang2021} we combine XAI and OSR by using a gradient-based method,  where gradient  backpropagation (limited to the last DL layer to keep computation simple)  is used to detect and explain large changes in the feed-forward model due that are due to the new input.
Output-based  OSR approaches have the advantage of a limited complexity and  direct explainability, as they  leverage direct model output, as opposite to surrogate models. Additionally, they are readily deployable, as they do not require modification of existing models --  which makes them particularly relevant and appealing for network AIOps.

\begin{figure*}[!t]
    \centering
 \includegraphics[width=0.46\textwidth]{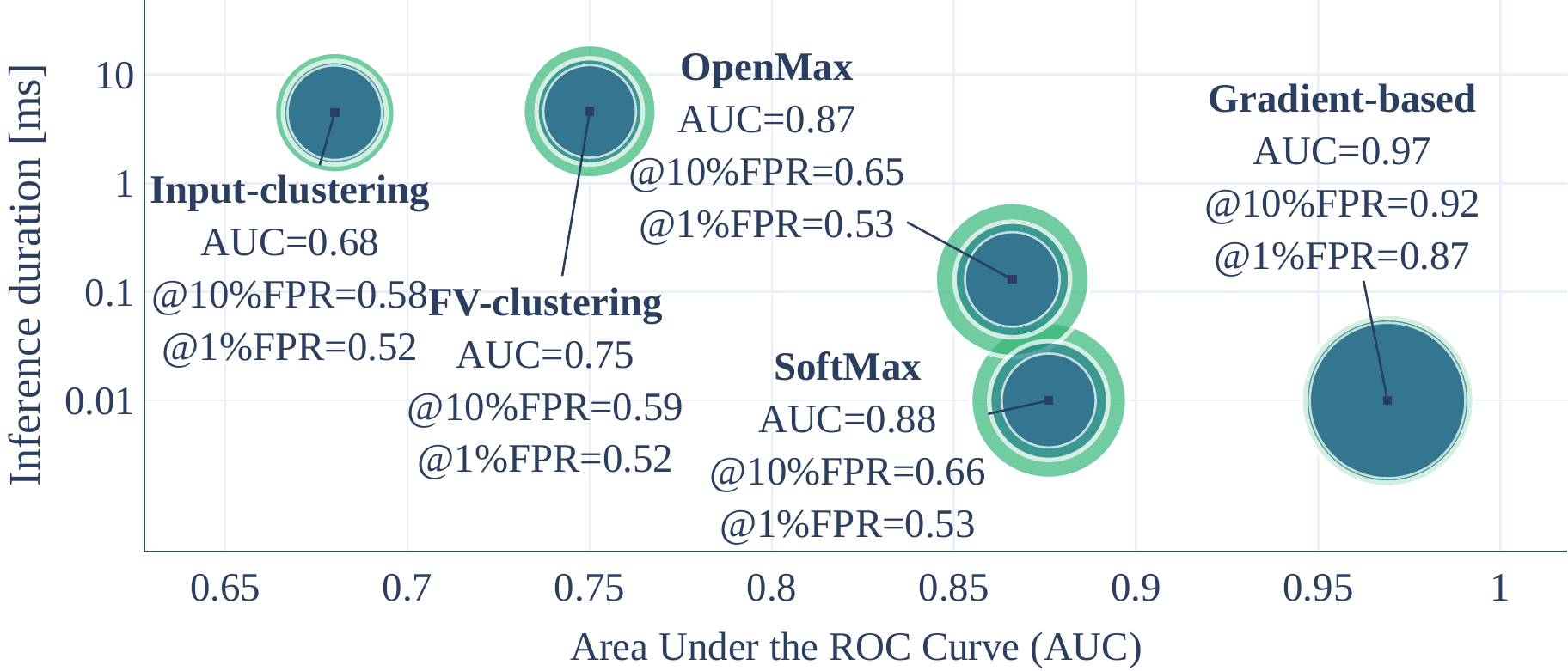}
 \includegraphics[width=0.46\textwidth]{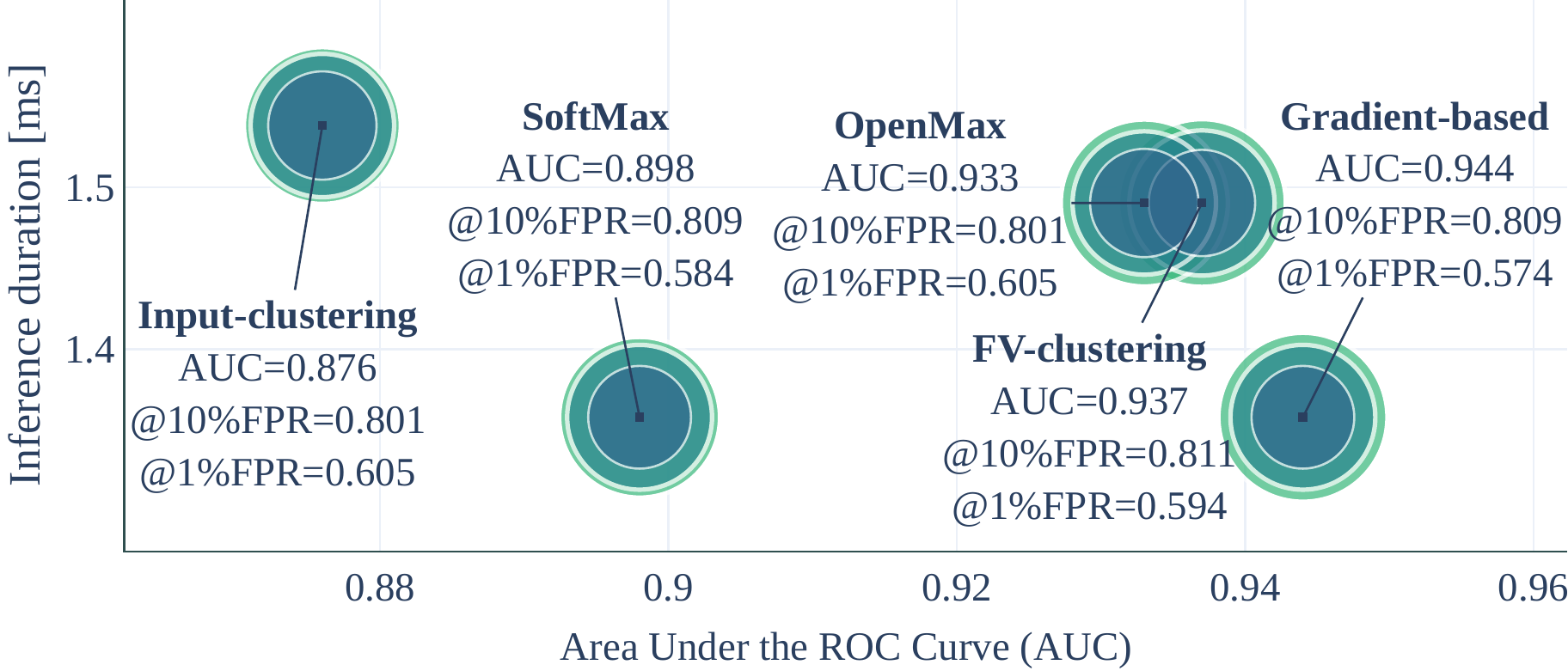} 
    \caption{\textit{Individual inference:} Performance-complexity scatter plot of several OSR techniques for  traffic classification  (left) and image recognition (right). }
    \label{fig:individual}
\end{figure*}

\section{Example AIOps use-cases }\label{sec:use-case}

We apply OSR for automated quality assessment of deployed DL models in two complementary use cases, related to \emph{network traffic classification} and \emph{image recognition}, that we compactly summarize and contrast in Table~\ref{tab:dataset}.

\begin{table}[!t]
  \begin{center}
    \caption{Use-cases at a glance. } \label{tab:dataset}
    \begin{tabular}{lll}
              &  {\bf Traffic  classification}  &
                 {\bf Image  recognition}  \\
    \toprule
      {\bf Dataset}  & Huawei (Proprietary)   & MNIST (Public)\\
      {\bf Classes}  & 835 applications       & 10 digits  \\
      {\bf Labels}   &  4.4M labeled flows    &  60k labeled images \\
    \midrule
      {\bf Known K}    & 200 apps, 4.5M flows   & 6 digits, 42K images \\
      {\bf Unknown U}  & 635 apps, 37K flows    & 4 digits, 28K images  \\
       {\bf    U/(U+K)\%}     & 76\% apps, 0.8\% flows & 40\% digits, 46\% images\\
    \midrule
      {\bf CNN model}  & 3 Conv$_{3\times1}$ + FC$_{128}$   &      2 Conv$_{5\times5}$ + FC$_{400}$ + FC$_{84}$     \\
      \bottomrule
    \end{tabular} 
    \end{center}
\vspace{-15pt}
\end{table}

\subsection{Use-case definition}

\subsubsection{Network traffic classification}




We consider the case of encrypted traffic classification, that is actively investigated  in the networking community nowadays~\cite{9146411}.
In contrast to identification of known applications, which is a well investigated subject and for which classic supervised methods are well suited, the network community has only very  limitedly \cite{Zhang15,carela2016streaming} dealt with handling applications that were never presented to the model during training, an OSR problem that is referred to as ``zero-day application'' detection in this context. In particular, the current state of the art \cite{Zhang15} performs k-means clustering on unmodified input, and is thus worth contrasting to data-science OSR solutions~\cite{Bendale15,Hendrycks2017,Zhang20,Yang2021}. A complementary approach is instead taken in \cite{carela2016streaming}, which does not tackle zero-day detection, but assumes a continuous stream of labels and  incrementally trains model to combate concept drift of known classes: OSR in this case would help detecting suspicious labels, or explain which classes are responsible for the most significant model changes.   

As baseline model, we use a 1d- Convolutional Neural Network (CNN) acting on the size and direction of the first packets of a flow, that is consistently found in the literature to have good accuracy for the identification of known applications~\cite{9146411,Yang2021}, and train the model to recognize the top-200 applications. To test OSR performance, we additionally fed the trained CNN model with 37k samples from 635 zero-day applications that were never presented at training. As typical in network traffic skew, the wide majority of samples correspond to few popular applications, while a long tail of zero-day application (76\%) account for only few flows ($<$1\%).

\subsubsection{Image recognition}
The traffic classification use-case represents a specific evaluation very relevant for the network O\&M field, but is conducted on proprietary dataset. We thus additionally consider an image-related use-case, which (i) is performed on public datasets thus \emph{reproducible}; additionally, (ii) image recognition is a \emph{network application}, when edge devices stream images to the cloud for processing, where novelty discovery is a tool to assess the fitness of the model to environment; finally, (iii) the use of image recognition allows to \emph{appreciate generality} of the methods. In particular, the MNIST handwritten digits and characters dataset is consistently used in the OSR literature~\cite{Bendale15,Hendrycks2017,Zhang20} as a benchmark.  In this case, the problem is for any input, to assess whether is part of the handwritten digits and characters shown to the model during training, or if is an entirely new symbol.

 We use LeNet~\cite{LeCunBoserDenkerEtAl89}, a classic 2d-CNN architecture  often used as a classification benchmark. Notice that CNN architectural choices  are qualitatively similar in both use-cases, although the  hyper-parameter tuning  (e.g., filter size and depth, layer size, etc.) quantitatively differ. In this case, we train the model to recognize 6 digits, and assess OSR ability to detect the other 4 digits, with a more balanced fraction of unknown labels (40\%) and samples (46\%).

\subsection{Experimental results}\label{sec:use-case:result}
We assess (i) how well and at which cost OSR techniques can detect novelty on \emph{individual samples,} as well as (ii) how to combine detection of novelty for individual inferences to allow quality assessment of an \emph{overall model}. 

\subsubsection{Individual inference}\label{sec:use-case:individual}
At high level, the OSR output is used to detect novel input never presented to the CNN model at training: i.e, OSR invalidates an
individual inference of the CNN model,  overriding an erroneous  known  label with a zero-day label. It is however possible that the OSR
technique is wrong (e.g., the app is not a zero-day) so that a correct CNN inference is wasted: the performance of the  OSR methods is thus well represented by the Area under the Receiver Operating Characteristic Curve (AUC). AUC  counts both correct  zero-day identification as well as false positive identification (and consequent CNN inference waste) and takes values in [0,1] with 1 indicating perfect OSR detection.

 Fig.\ref{fig:individual} reports a performance-complexity  scatter plot of the different OSR techniques for both use-cases: best performance are for techniques falling in the bottom right portion of the plot.
 The outer circle represent the overall AUC, whereas inner circles report the AUC value for a maximum waste (i.e., 1\% and 10\%  false positive rate). 
  As the 1d-traffic classification dataset is much larger (in classes) and skewed (in samples per class) than the 2d-image,  the different use-cases  let us contrast  complementary OSR  operational points.
  
  In terms of (i) \emph{accuracy}, for the 1-dimensional time-series problem of the traffic classification use-case, OSR techniques are well dispersed:  gradient-based\cite{Yang2021} performs the best (significantly better than SoftMax\cite{Hendrycks2017} and OpenMax\cite{Bendale15}, with very similar complexity) and  input-clustering~\cite{Zhang15}  the worst (due to the ``curse of dimensionality'' and the high number of clusters, over which  output-clustering~\cite{Zhang20} only provides a very limited advantage). In the bi-dimensional image use-case, performance of top-methods are in a closer range, but the gradient-based method still consistently shows superior performance.

 In term of (ii) \emph{time-complexity} for the traffic classification use-case, notice that SoftMax and Gradient techniques allow in excess of 100,000 OSR zero-day operations per second, and are thus amenable to assess the quality of the model at \emph{each inference} -- i.e., they can thus be used for exhaustive tracking. Conversely, clustering techniques\footnote{Since the number of clusters is proportional to the number of known classes, the complexity of the method is more apparent for traffic classification than for MNIST, where the number of classes is 20$\times$ smaller.} only allow about 100 OSR operations per second, so that they would require to \emph{significantly down-sample} (by a factor of 1000$\times$) inference results to be validated by  OSR -- in turn possibly limiting the   model validation timeliness.

 \begin{figure}[!t]
    \centering
 \includegraphics[width=0.46\textwidth]{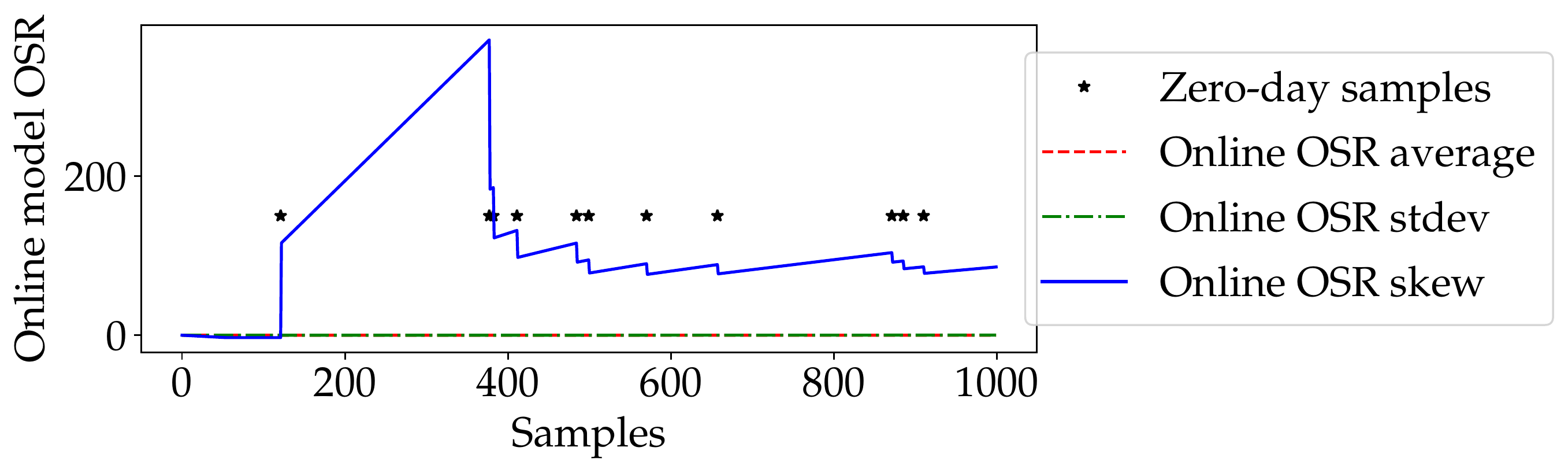}\\
 \includegraphics[width=0.46\textwidth]{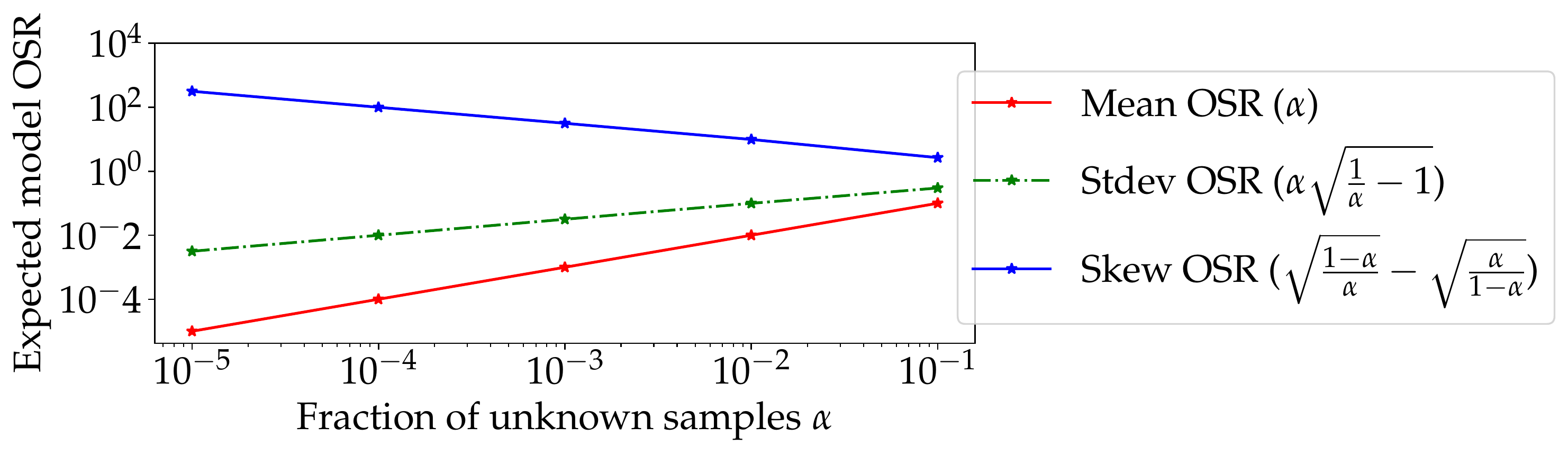} 
    \caption{\textit{Overall model quality tracking}: online OSR time evolution (top) and expect OSR moments for different zero-day rates (bottom).}
     \label{fig:overall}
\vspace{-4ex}
\end{figure}

\subsubsection{Tracking overall model quality}\label{sec:use-case:tracking}
 We now show  that by tracking OSR output of individual inferences (e.g., with SoftMax, OpenMax or Gradient methods), it is possible to robustly assess the (i) \emph{overall quality} of a model (e.g., by simple online average of OSR over all samples), as well as (ii) \emph{timely track} subtle changes in the environment (e.g., from the very first unknown samples, by tracking higher moments of the OSR results). 
 We stress that, in the traffic classification use-case, the bulk of popular applications constitute the majority of samples. As the fraction $\alpha$ of zero-day samples is  rare ($\alpha < 1\%$),
  the first unknown samples are going to be  presented to the DL models after several thousands  of known inputs (or even more for use-cases with further class imbalance, or DL models with better class coverage).
  
  
  
  We illustrate model quality tracking  in  Fig.\ref{fig:overall}.
  The top portion shows  an excerpt of the zero-day arrival process for a batch of 1,000 flows, along with the unbiased online estimators of the OSR output average, standard deviation and skew: changes in the OSR skew are clearly visible at each new zero-day sample arrival (black star).
  The bottom  picture of \Cref{fig:overall} further reports the expected moments of the OSR statistics, for a fraction of unknown samples equal to $\alpha$
  (for traffic classification, $\alpha$$\approx$$10^{-4}$). It can be seen that the mean model-level OSR increases as a function of the zero-day rate $\alpha$ (more precisely equals  $\alpha$):  as such, the (i) \emph{average} is suited to express \emph{significant degradation of the model quality}, but is not suited to detect rare events.  Conversely, model-level OSR skew grows for decreasing $\alpha$ (precisely as $\sqrt{\frac{1-\alpha}{\alpha}}-\sqrt{\frac{\alpha}{1-\alpha}}$): as such, the (ii) \emph{skew} is well suited to  especially appreciate  rare events and can be leveraged as a \emph{reliable and timely indicator of early model degradation}.



\section{Conclusion and discussion}\label{sec:conclusion}
This article discusses AIOps, illustrating how  data science best practices and techniques can be shared across all Dev, Ops and Biz stakeholders. In particular, 
it identifies in Open-set recognition (OSR) a key building block for 
(i) judging the fit of a trained model to its deployment environment, by detecting novelty in individual inferences, as well as for  
(ii) timely and accurate detection of overall model degradation, by tracking multiple inferences of the same model. 
We propose simple, effective and flexible techniques that we contrast with the state of the art, on two radically different use cases -- namely, traffic classification, and image recognition: experimental results testify the feasibility of general, autonomous, accurate and timely quality assessment of deployed DL models. 

Clearly, more research is needed to further automate  and simplify the transfer of data science best practices, concerning both (i) the narrow subfield of OSR that we focus on in this paper, as well as (ii) the broader set of techniques that pertain to the full-blown  AIOps loop.   For instance, OSR techniques could benefit from Deep Bayesian frameworks, which are  better suited to model the in-distribution data: at the same time, inference 
time will remain a  critical factor for deployment,  and should not be neglected by future  work.  Finally, a larger set of techniques complementary to OSR is needed across other AIOps steps: for instance, while OSR can assist to  timely trigger (and prioritize) model retraining, however efficient techniques for incremental learning (e.g., add new class or domain adaptation of existing classes) and decremental unlearning (e.g., remove old class) are going to be crucial in simplifying model updated; similarly, while OSR can be helpful in detecting mislabeling, techniques to cope with label scarcity (e.g., self-supervised, few-shot learning) will alleviate the cost of retraining.  Collectively, this enhanced set of automated machine learning tools will enrich and streamline AIOps development and operation.

\section*{Acknowledgements}
The authors wish to thank the Associate Editor and the Reviewers for whose useful feedback helped improving the quality of this paper.

\begin{small}
\bibliographystyle{IEEEtran}
\bibliography{references}
\end{small}

\balance 

\begin{IEEEbiography}[{\includegraphics[width=1in,height=1.25in,clip,keepaspectratio]{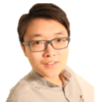}}]{Lixuan Yang}
is a Senior Engineer at the DataCom Lab of Huawei’s Paris Research Center. She received her Ph.D. from the CNAM (2017) and MSc from Jean Monet University (2013). Her current interests include federated learning, novelty discovery and continual learning applied on traffic classification.
\end{IEEEbiography}

 \begin{IEEEbiography}[{\includegraphics[width=1in,height=1.25in,clip,keepaspectratio]{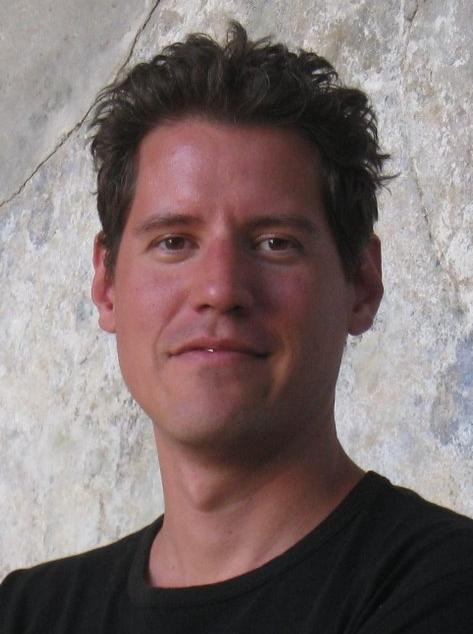}}]{Dario Rossi} is  Director of the DataCom Lab at Huawei Technologies, France. Previously held Full Professor positions at Telecom Paris and Ecole Polytechnique. He coauthored 15+ patents and 200+ papers, and  received 9 best paper awards, a Google Faculty Research Award and an IRTF Applied Network Research Prize. 
\end{IEEEbiography}




\label{last-page}

\end{document}